\documentclass[preprint,12pt]{elsarticle}

\usepackage{graphicx}%
\usepackage{multirow}%
\usepackage[english]{babel}
\usepackage{amsmath,amssymb,amsfonts}%
\usepackage{amsthm}%
\usepackage{mathrsfs}%
\usepackage[title]{appendix}%
\usepackage{xcolor}%
\usepackage{textcomp}%
\usepackage{manyfoot}%
\usepackage{booktabs}%
\usepackage{algorithm}%
\usepackage{algorithmicx}%
\usepackage{algpseudocode}%
\usepackage{listings}%
\usepackage{lmodern}
\usepackage{anyfontsize}
\usepackage{enumitem}

\usepackage[colorlinks=true,linkcolor=blue]{hyperref}

\journal{Swarm and Evolutionary Computation}
 \newenvironment{changed}{} {}

\begin{document}

\begin{frontmatter}

\title{The Paradox of Success in Evolutionary and Bioinspired Optimization: Revisiting Critical Issues, Key Studies, and Methodological Pathways}

\affiliation[label1]{organization={Department of Computer Science and Artificial Intelligence, Andalusian Research Institute in Data Science and Computational Intelligence (DaSCI)},
            city={Granada},
            postcode={18071},
             country={Spain}}

\affiliation[label2]{organization={TECNALIA, Basque Research \& Technology Alliance (BRTA)},city={Derio},postcode={48160},country={Spain}}

\affiliation[label3]{organization={Department of Mathematics, University of the Basque Country (UPV/EHU)},city={Leioa},postcode={48940},country={Spain}}

\author[label1]{\corref{cor1}{Daniel Molina}}
\ead{dmolina@decsai.ugr.es}
\author[label2,label3]{Javier {Del Ser}}
\ead{javier.delser@tecnalia.com}
\author[label1]{Javier Poyatos}
\ead{poyatosamador@gmail.com}
\author[label1]{Francisco Herrera}
\ead{herrera@decsai.ugr.es}

\cortext[cor1]{Corresponding author}

\begin{abstract}
Evolutionary and bioinspired computation are crucial for efficiently addressing complex optimization problems across diverse application domains. By mimicking processes observed in nature, like evolution itself, these algorithms offer innovative solutions beyond the reach of traditional optimization methods. They excel at finding near-optimal solutions in large, complex search spaces, making them invaluable in numerous fields. However, both areas are plagued by challenges at their core, including inadequate benchmarking, problem-specific overfitting, insufficient theoretical grounding, and superfluous proposals justified only by their biological metaphor. This overview recapitulates and analyzes in depth the criticisms concerning the lack of innovation and rigor in experimental studies within the field. To this end, we examine the judgmental positions of the existing literature in an informed attempt to guide the research community toward directions of solid contribution and advancement in these areas. We summarize guidelines for the design of evolutionary and bioinspired optimizers, the development of experimental comparisons, and the derivation of novel proposals that take a step further in the field. We provide a brief note on automating the process of creating these algorithms, which may help align metaheuristic optimization research with its primary objective (solving real-world problems), provided that our identified pathways are followed. Our conclusions underscore the need for a sustained push towards innovation and the enforcement of methodological rigor in prospective studies to fully realize the potential of these advanced computational techniques.
\end{abstract}


\end{frontmatter}

\section{Introduction}\label{sec:introduction}

Bioinspired computation, a prominent area within metaheuristic optimization research \cite{Yang2013}, focuses on the design of computational techniques and optimization algorithms that draw inspiration from biological processes and behaviors observed in nature. This inspiration allows such algorithms to solve complex optimization problems in a more efficient and robust manner than traditional solvers. Evolutionary computation \cite{Back1997} plays a central role in this research area, finding its inspiration in principles from Darwin's Theory of Evolution to tackle intricate optimization problems that often surpass the capabilities of traditional approaches \cite{Kar2016,Yao1999,Ashlock2006}. Both fields have seen significant practical advancements \cite{del2019bio}, demonstrating their potential to improve other models, from classic Machine Learning (ML) pipelines \cite{al-sahaf2019a} to deep neural networks \cite{zhan2022,martinez2021lights}.

Recent studies have anticipated the optimization challenges posed by the advent of General-Purpose Artificial Intelligence Systems (GPAIS) \cite{triguero2024b}, in which the flexibility of bioinspired optimization approaches can address the complexity of GPAIS for several learning tasks. Examples include Generative Adversarial Networks in zero-shot learning \cite{Chen2024}, prompt evolution for Large Language Models \cite{Guo2024}, and evolutionary algorithms for transfer learning and pruning in deep neural networks \cite{Poyatos2023}.

The surge in publications that present new bioinspired optimization algorithms has motivated efforts to systematically classify this vast body of literature in different taxonomies \cite{Stork2020,Rajwar2023}, as well as comprehensive overviews of the topic \cite{Ferrer2023,Marti2024,Velasco2024}. Several of these contributions have been continuously updated over time, leading to more sophisticated taxonomies that incorporate descriptions of novel overviews and methodologies. A notable example is \cite{Molina2020}, whose latest version maintained in \cite{Molina2020arxiv} has systematically collected, analyzed, and classified more than 500 bioinspired solvers to date. 

Although the explosive growth and success of bioinspired computation have established it as a cornerstone of metaheuristics, they have ironically given rise to a \emph{paradox of success}. The influx of academic interest and sustained flow of contributions have not only accelerated advancements, but also increased the concerns of part of the community around critical issues and bad practices noticed in the area. Such concerns were first manifested in the seminal work by S{\o}rensen in \cite{Sorensen2015}. This work, along with other contemporary manifests focused on criticizing specific solvers, was the first of its kind to expose that a metaphor is in no way a factor guaranteeing the novelty of a bioinspired solver. Since then, the number of critiques of such practices has increased sharply, highlighting the lack of novelty of certain algorithms or exposing poor methodological practices in benchmarking and evaluation \cite{Tzanetos2021,Campelo2023}. Remarkably, the importance of this growing corpus of critiques has motivated the publication of an open letter formally rejecting noninnovative proposals, signed by front-line experts in this research area \cite{Aranha2022}.

On a positive note, the research community is addressing shortcomings in bio-inspired algorithms by proposing methodological guidelines, quantifying differences between algorithms and their operators \cite{Hu2024}, establishing rigorous comparison guidelines \cite{LATORRE2021}, automating algorithm design \cite{Camacho2023}, and developing appropriate benchmarks \cite{Kumar2021}. These initiatives aim to promote best practices and drive transformation in the field, ensuring that bioinspired and evolutionary computation continue to advance robustly in years to come.

In this context, this manuscript presents an overview of the weaknesses and criticisms that have appeared in the literature over the years, delving into the rationale given by the authors for their claims. In addition, it also analyzes the different methodological pathways that have been put forward to address these criticisms and to improve the field. By examining the two sides of the coin, the problems and the corresponding solutions, our study provides a comprehensive analysis of the aforementioned \emph{paradox of success}: a plethora of publications presenting new solvers and/or applications using bioinspired or evolutionary algorithms, but unfortunately flooded by algorithmic designs do not really offering a significant contribution to the area, mainly due to a lack of novelty beyond the biological metaphor inspiring its design. As will be discussed later, it is time to separate the wheat from the chaff, starting from a positive recapitulation of the current state of the area in this regard, and ending with referential guidelines to drive efforts in the field toward impactful scientific advancements and practical applications. This recapitulation is indeed the overarching goal of this work.

The paper is structured according to this overall goal. First, Section~\ref{sec:critical} addresses the problems, examining critical issues known in the bioinspired and evolutionary computation fields and the key studies that expose them. Next, Sections~\ref{sec:good}, \ref{sec:frc} and \ref{sec:methodologies_improving} showcase the efforts made by the community to address these critical problems through the development of proper methodological pathways for the benefit and progress in the area. Next, Section \ref{sec:automation_design} focuses on the automated design of metaheuristics algorithms as a research area that can overcome the shortcomings discussed previously. Finally, Section~\ref{sec:conclusions} draws the main conclusions from this study. 

\section{Known Critical Issues: From Lack of Innovation to Low-Quality Experimental Studies} \label{sec:critical}

As mentioned in the introduction, the number of bioinspired solvers has grown significantly in recent years,
owing to methods that draw inspiration from natural behaviors or processes. However, many proposals lack innovation and experimental rigor, hindering the field's progress.
It is paradoxical that the field's success in terms of publications and proposals is what threatens its real progress. In agreement with the title of this paper, this section examines the issues leading to this \emph{paradox of success}: the lack of algorithmic innovation (Section \ref{sec:lack}), the neglect of critical aspects of experimentation (Section \ref{sec:poor_exp_studies}), and the failure to use appropriate benchmarks (Section \ref{sec:poor-benchmarks}).


\subsection{On the Lack of Algorithmic Innovation} \label{sec:lack}


The plethora of bioinspired algorithms available poses a significant challenge in choosing the best solver for an optimization problem.
In this context, the work of Molina et al.~\cite{Molina2020} proposes a dual taxonomy according to their inspiration and algorithmic behavior. They highlight that: 
\begin{itemize}[leftmargin=*]
    \item []
``\textit{In summary, although in the last years many nature-inspired algorithms have been proposed by the community and their number grows steadily every year, more than half of the proposals reviewed in our work are incremental, minor versions of only three very classical algorithms (Particle Swarm Optimization, Differential Evolution, and Genetic Algorithms). We therefore conclude that the large number of natural and biological sources of inspiration used so far to justify the design of new optimization solvers has not led to significantly disruptive algorithmic behaviors}''. 
\end{itemize}

\noindent In other words, many recently developed algorithms, inspired by natural phenomena or processes/behaviors observed in other fields of knowledge, frequently recycle existing ideas without introducing algorithmically new methods or approaches.

This has been a concern for almost a decade. As was discussed in \cite{Sorensen2015}, the fallacies of ``novel'' metaphor-based methods and the vulnerability of bioinspired research emphasized the need to impose good research practices in the area. In the author's own words, ``\textit{most ``novel'' metaheuristics based on a new metaphor take the field of metaheuristics a step backward rather than forward}. That study is a ``\textit{call for a more critical evaluation of such methods}'' because ``\textit{there are more than enough novel methods, and there is no need for the introduction of new metaphors just for the sake of it}''.

Before Sorensen's manifesto was published, other researchers \cite{weyland2010,Piotrowski2014} had also addressed the lack of innovation of specific bioinspired solvers, claiming that the diversity of their natural inspirations did not extend to mathematical differences with previous evolutionary algorithms. \citet{weyland2010} (together with a series of follow-up rebuttals and responses \cite{geem2012research,weyland2015critical}) provided a comprehensive review of \emph{Harmony Search}, and critically analyzed its developments over the past decade. He argued that this bioinspired solver, often seen as a significant innovation, is essentially a special case of Evolution Strategies. The second paper \cite{Piotrowski2014} focused on a bioinspired approach, called \emph{Black Hole Optimization}, demonstrating that it is, in fact, a ``\textit{simplification of the well-known Particle Swarm Optimization with inertia weight}''. Furthermore, the authors express their concern about the fact ``\textit{that such an approach may jeopardize achievements in the field of Evolutionary Computing}'', since Black Hole Optimization is a simplification that produces favorable results in specific benchmark tests.

Later, the work in \cite{Fister2016} discussed this phenomenon in relation to the number of papers published, arguing that ``\textit{the new population-based nature-inspired algorithms are released every month and, basically, they have nothing special and no novel features for science}''. Their conclusions are aligned with the taxonomy published in \cite{Molina2020}, because ``\textit{our research revealed that the process of the new population-based nature-inspired algorithm possesses the behavior of the swarm intelligence paradigm}'', revealing that the category with the highest number of contributions is Swarm Intelligence.


Thereafter, in \cite{Tzanetos2021} a meticulous review of the trends of new algorithms was performed, providing additional criticisms that apply to most of these proposals:
\begin{enumerate}[leftmargin=*]
\item The existence of a physical analogy, as there are some examples of bioinspired algorithms that present a phenomenon that does not exist in nature.
\item The similar inspiration and even duplicate methods, due to the existence of groups of algorithms very similar to established solvers, with subtle differences lying on slight modifications to their algorithmic components.
\item Multiple techniques drawing inspiration from the same idea and with authors publishing multiple variants over time under this premise. Interestingly, this observed practice escalates to research groups publishing ``novel'' algorithms that solve the same set of functions and problems.
\item New natural-inspired algorithms provide added value when their contribution is crucial. There are some cases in which the introduction of novel algorithms is necessary, and in those cases, the development of novel algorithms is considered a suitable solution. 
\end{enumerate}

In addition, the study in \cite{Tzanetos2021} examines the circumstances in which it is necessary to develop a new bioinspired algorithm. Two particular cases merit further attention, as they exemplify the underlying postulates that drive our analysis: 
\begin{itemize}[leftmargin=*]
\item ``\textit{when it manages to solve a problem in a way superior to the way it was solved by previous competitive approaches}''; or when 
\item ``\textit{more intelligent mechanism has been incorporated within this new nature-inspired algorithm that seems to render it more efficient than others}''. 
\end{itemize}

They agree on the importance of its application to real-world problems and highlight the fact that ``\textit{the main goal from now on should be the attempt to upgrade and fine-tune the existing nature-inspired algorithms, to obtain competitive solutions in real-world applications}''.

The considerable number of proposals that fail to introduce genuinely new concepts
not only fail to contribute to the field, but also present a significant challenge for researchers trying to stay up-to-date, due to the difficulty of identifying relevant proposals among the plethora of irrelevant proposals. Fortunately, several papers have attempted to identify non-innovative proposals, which we will describe in Section \ref{sec:identify_bad_proposals}.

Although it is beneficial to have resources to identify ineffective proposals, this is not a sufficient solution. It is essential to understand the underlying factors that contribute to the exponential growth of the number of proposals. In this context, \citet{Campelo2023} remark that a possible cause is the pressure to \textit{publish or perish}, which is perceived as a low-effort, low-risk process with high potential rewards. Some authors have built professional careers from developing an upsurge of metaphor-based methods. The authors of this paper maintain a regularly updated collection of bioinspired proposals in the well-known Evolutionary Computation Bestiary\footnote{Evolutionary Computation Bestiary, \url{http://fcampelo.github.io/EC-Bestiary/}, accessed on January 10th, 2025.}. In their words, ``\textit{the phenomenon of metaphor-inspired methods has had a negative impact on the field, wasting the work of scientists and reviewers on methods that continuously reinvent the wheel, hiding sloppy or dubious practices}''. In this context, the aforementioned \textit{publish or perish} phenomenon has been taken to new extremes, all in an effort to gain more influence in the field. As described by the authors, ``\textit{although the critical tone of the Bestiary is clearly stated in the repository, we are often contacted by authors of ``novel'' metaphor-based metaheuristics requesting that their work be listed. It has never been clear to us whether these authors do not understand the critical tone of the page or if they assume that any exposition, however critical, would be a net positive for their work}''.

For this reason, and to avoid falling into this bad practice, in \cite{Aranha2022} the authors wrote a letter to ``calling on all editors in chief in the field to adapt their editorial policies'' to reject the publication of \emph{allegedly novel} metaphor-based bioinspired algorithms.
This letter underscores that the authors of new proposals have succeeded in publishing them in journals related to the metaphor itself, without a comprehensive and principled editorial assessment of the computational advantages and benefits of the presented algorithm: ``\textit{For instance, if the metaphor is the mating behavior of bats, the authors will attempt to publish it in specialized journals on bats or animal mating behavior; metaheuristics inspired by improvising musicians will go for journals on music}''. For these reasons, almost a hundred important researchers in the area agreed to sign this letter and accept the publication of novel bioinspired algorithms under the following postulates:
\begin{enumerate}[leftmargin=*]
\item \textit{``present their method using the normal, standard optimization terminology''}; 
\item \textit{``show that the new method brings useful and novel concepts to the field''}; 
\item \textit{``motivate the use of the metaphor on a sound, scientific basis''}; and 
\item \textit{``present a fair comparison with other state-of-the-art methods using state-of-the-art practices for benchmarking algorithms''}.
\end{enumerate}

Despite the critical manifests expressed by several authors over the years, concerns regarding these issues continue to persist today. This is evinced by several works published during this year. For example, the study in \cite{Jiao2024} examines various natural mechanisms and certain ``\textit{commonalities of nature-inspired intelligent computing paradigms. These commonalities provide a solid algorithmic foundation to avoid designing unreasonable metaphors}''.


Another work in the same line is \cite{Deng2024}, that analyzes various bioinspired algorithms, focusing on Arithmetic Optimization Algorithm's mathematical formulation, implementation, and lack of unbiased comparisons. The authors conclude that ``\textit{the current trajectory of focusing on novelty over theoretical foundations jeopardizes the credibility and long-term sustainability of advancements in the field. Hence, it becomes imperative for researchers to direct their efforts toward developing and substantiating theoretical foundations for metaheuristic algorithms}''. Another study that exposes the lack of novelty of a specific modern bioinspired solver is \cite{Halsema2024}, which performs a comparison of the algorithm called \emph{Raven Roost Optimization}. Their analysis reaches a conclusion similar to that of previous studies dealing with other solvers. Raven Roost Optimization ``\textit{contains no real novelty compared to the well-established Particle Swarm Optimization, and in fact, will perform worse in many situations due to the inherent bias towards its starting point}''. 

We consider that the combination of critical analysis of existing proposals, in conjunction with stricter criteria for the acceptance in the literature of new proposals, giving special attention to their contribution, could address the lack of algorithm innovation.  We provide further details in this regard in Section \ref{sec:good}.


\subsection{On the Low Quality of Experimental Studies} \label{sec:poor_exp_studies}

Another significant concern in bioinspired and evolutionary computation is the quality of the experimental benchmarks. A substantial number of proposals in the literature conduct insufficient comparative studies to assess the performance of the developed algorithms. The primary issue is their failure to compare with the state-of-the-art methods and similar solvers. To demonstrate any improvements, it is essential to perform comparisons against competitive algorithms and validate that the individual components of the proposals genuinely contribute to enhancing overall performance.
However, this is not the case in many of such studies. as underscored in \cite{Molina2020}:
\begin{itemize}[leftmargin=*]
    \item []
``\textit{when new algorithms are proposed, unfortunately, many of them are only compared to very basic and classical algorithms (such as Genetic Algorithms or Particle Swarm Optimization). These algorithms have been widely surpassed by more advanced versions over the years, so obtaining better performance than the naive version of classical algorithms is relatively easy to achieve, and it does not imply competitive performance}''. 
\end{itemize}

Furthermore, the set of competitive algorithms should be chosen considering the state-of-the-art of each benchmark \cite{Molina2018a}, as the behavior of different bioinspired and evolutionary algorithms can largely depend on the characteristics of the benchmark under consideration
\cite{molinaAnalysisWinnersDifferent2017a}. In this regard, a recent study by \citet{piotrowskiChoiceBenchmarkOptimization2023c} concludes that ``\textit{in the present paper we show that the choice of the set of benchmarks used for the comparison may greatly affect the ranking of optimizers}''.

Another common criticism expressed against many modern bioinspired algorithms is that the algorithm itself is too complex. A reduction in the number of algorithmic components would facilitate a more comprehensive understanding and analysis of the algorithm's constituent parts. In this line of reasoning, the extensive experiments carried out in \cite{Piotrowski2018} showed that ``\textit{some meta-heuristics have to be simplified because they contain operators that structurally bias their search by favoring sampling from some parts of the decision space}''. Moreover, this study analyzed the L-SHADE-EpSin algorithm, which is a step-by-step designed algorithm developed from different variants of Differential Evolution. In their experimentation, the authors proved that ``\textit{the proposed simplified variant of L-SHADE-EpSin is highly competitive in a wide collection of artificial benchmarks and real-world problems}'' compared to other bioinspired algorithms and the non-simplified version of this algorithm.

\subsection{On the Use of Poor Benchmarks}
\label{sec:poor-benchmarks}

Many studies lack an appropriate and unbiased benchmark, using their own comparison functions, which limits the generalizability of their findings to real world problems.  This practice may introduce biases that distort the accurate assessment of the performance of any new optimization algorithm, leading to misleading conclusions.

\begin{changed}
In this context, the work in \cite{rajwarStructuralBiasMetaheuristic2025} examines the different types of biases present in common metaheuristic algorithms. Among them, the study concludes that origin bias and center bias are the most recurrent ones, due to the
\end{changed}
center bias operators that appear in some bioinspired and evolutionary algorithms. This phenomenon has been thoroughly studied in \cite{Kudela2022} by systematically analyzing this behavior between several bioinspired algorithms. The reported experiments show a significant tendency in most modern bioinspired solvers to explore mainly the center of domain search. Thus, the majority of these algorithms generate misleading performance results because they are compared considering objective functions whose global optima are located at the center of the solution space. Most of these biases can be avoided by resorting to specific benchmarks proposed over these years, as explained in Section \ref{sec:frc}. 

One reflection we cannot lose sight of within this matter is that benchmark performance cannot be a goal in itself. Benchmarks are a valuable tool for the design and evaluation of algorithms that can potentially perform well in real-world problems. Similarly to \cite{Aranha2022}, \citet{Ceberio2024} discuss \textit{``the role of experimentation in the two approaches of conducting research: engineering vs. scientific''}, with thoughtful remarks about benchmarks and experimentation details, and present their understanding of the fundamental principles of both approaches. For this disparity, Kudela proposes the incorporation of a greater number of real world problems into the benchmark \cite{Kudela2022}, by ``\textit{the construction of a large set of challenging real-world benchmark problems}''. In this way, the author advocates for the establishment of a repository that includes the following:
\begin{enumerate}[leftmargin=*]
\item \textit{``several heterogeneous benchmark sets with a unified way of calling the test problems''}; 
\item \textit{``trusted implementations (source codes) of both standard EC methods and up-to-date state-of-the-art techniques''}; 
\item \textit{``data obtained from running the algorithms (from (2)) on the benchmarks (from (1))''}.
\end{enumerate}

Alternatively, another methodological path that can be followed when benchmarking bioinspired solvers is to continue using synthetic functions. However, as argued in \cite{vanderblom2023}, ``\textit{the design of benchmarks would benefit from a better understanding of which properties appear in real-world problems, how common these properties are and in which combinations they are found. In addition, it is of particular interest to identify characteristics of real-world problems that are not yet represented in artificial benchmarks and to identify real-world problems that might be usable as part of a benchmark suite. All these aspects would then provide better guidance to the development of algorithms that perform well in the real world}''.

Therefore, it becomes crucial to assess the characteristics of the benchmarks in use, as done in \cite{Hellwig2019,mersmann2010,yin2024}, and to analyze whether these characteristics match those of real world problems. An interesting work on this matter is \cite{vanderblom2023}, in which, through a questionnaire, several important properties are identified, such as the presence of noise, solution constraints, multiple but reduced objectives, or unknown optima, among others. Such characteristics are not always effectively represented in synthetic benchmarks. 

Here, comes a major cornerstone of benchmarking in metaheuristic optimization: their \emph{representativeness}. It refers to the extent to which a benchmark function accurately reflects the characteristics and challenges of real-world optimization problems.
A \emph{representative} benchmark ensures that the performance of metaheuristic algorithms is assessed in a way that is both meaningful and transferable to real-world scenarios.
In this regard, \citet{Chen2024benchmark} studies the importance of benchmark \textit{representativeness} in evaluating optimization algorithms. To this end, it defines three levels of this quality depending on the set of problems to be represented, and introduces a quantitative metric based on several features of benchmark problems, including separability, elementary functions, or dispersion metric, among others.

Another fundamental aspect that is sometimes ignored is to keep in mind that, for real-world problems, there are other qualities of the proposals, apart from the performance. These qualities, like their complexity, hardware requirements, for distributed models the achieved speedup and robustness, or the required time to obtain a reasonably good solution, can be even more relevant than the performance when choosing the right optimizer in many practical applications \cite{OSABA2021}.
\begin{changed}
Moreover, it is crucial to note that while the stopping criterion is predetermined in most benchmarks, in practical applications this aspect is often more flexible and subject to external constraints. Therefore, it is necessary to evaluate the metrics for a range of function evaluations to
gain a more comprehensive understanding of the algorithms' behavior in different scenarios \cite{piotrowskiMetaheuristicsShouldBe2025a}.
\end{changed}

All these experimental problems will be further examined in Section \ref{sec:frc} and illustrated with best practices methodologies for researchers in Section \ref{sec:methodologies_improving}. We believe that if these guidelines are applied to new proposals and adherence to them is encouraged by reviewers in editorial processes, these noted experimental issues can be effectively avoided.

\section{Separation of Wheat from Chaff: Distinction of Weak Proposals} \label{sec:good}


As noted previously, the field of bioinspired algorithms remains problematic due to malpractices and ineffective metaphorical designs, which often result in algorithms that provide no scientific value for the field, for lacking sufficient innovation or flawed experimental designs. We refer to these algorithms as \emph{weak proposals}.
In Section \ref{sec:identify_bad_proposals}, we describe several key studies that expose these \emph{weak proposals} that do not incorporate innovation. In Section \ref{sec:pathways_weak}, we propose several pathways to carry out studies to identify
these \emph{weak proposals} in the literature, or to avoid these problems when designing new proposals.

\subsection{On the Distinction of Weak Proposals} \label{sec:identify_bad_proposals}

Fortunately,  in recent years, several authors have recognized the need for a more rigorous development of new proposals along with a robust experimental framework, and identify in the literature several well-known proposals whose contribution in the field could be considered questionable, or \emph{weak proposals}.

Analyzing and identifying lack of novelty in bioinspired optimization algorithms has been a focus since early works (cf.\ \cite{weyland2010, Piotrowski2014}). These early studies built a foundation for evaluating particular solvers in the field, and their influence has persisted since their publication. An example is Teaching-Learning-Based-Optimization, which was analyzed in \cite{Pickard2016}. This study found that the algorithm has a bias towards the origin during teaching, which increases as the population converges. In addition, the Gravitational Search Algorithm was analyzed in \cite{Gauci2012}, finding that the force model in the algorithm depends only on agents' masses, not distances, contradicting the law of gravity.

Years after these studies, \citet{Camacho2020} provided evidence that the Grey Wolf, Firefly, and Bat Algorithms are not novel, but rather reformulations of ideas
initially introduced for Particle Swarm Optimization and subsequently refurnished. In order to prove this claim, the search operators of such solver were rewritten and mathematically compared, concluding that they can all be regarded as existing variants of the Particle Swarm Optimization algorithms. In the studies mentioned in \cite{Camacho2018, Camacho2019intelligent, Camacho2022}, several bioinspired algorithms, including the Intelligent Water Drops and Cuckoo algorithms, are critically examined. The authors argue that the metaphors used to justify these algorithms are often oversimplified or misapplied, leading to misleading or even deceptive claims regarding their novelty and effectiveness. 
Another example is \cite{Halsema2024}, analyzing the Raven Roost Optimization. The methodology of this paper involved a detailed examination of both the mathematical formulation and an analysis of the behavior, exposing a serious weakness (namely, an inherent search bias towards the starting point).


In the last couple of years, a notable increase in studies critically analyzing certain bioinspired algorithms by comparing them to others by examining their individual components or operators has been observed. \citet{Camacho2023b} contend that the well-known Grey Wolf, Moth-flame, Whale, Firefly, Bat, and Antlion algorithms are not actually novel. A comprehensive component-based analysis of each algorithm to substantiate this assertion, these algorithms are identified as variants of Particle Swarm Optimization and Evolution Strategies.
In this context, the work of \citet{Molina2020} categorizes various bioinspired algorithms based on their closest classical solver in terms of algorithmic structure. This analysis highlights that the distinctions between many bioinspired algorithms and their classical counterparts are often insufficient at the algorithmic level.


The Salp Swarm Optimization Algorithm is a further example of a bioinspired algorithm that has recently been subject to criticism. \citet{Castelli2022} identify several problems with this algorithm, which can be broadly classified according to its update rule, its physical motivation, and the discrepancy between their description and its available implementation. In addition, the study demonstrates that the original algorithm exhibits a bias towards the center, a point that was previously discussed. This bias toward the center was also detected in the Arithmetic Optimization Algorithm, analyzed in \cite{Deng2024} in terms of its mathematical formulation. As with other solvers, the lack of shifted versions of standard functions hinders the detection of this bias.
Whale Optimization Algorithm has recently entered this list of weak algorithms \citet{Deng2024Whale}, proving the existence of a center-bias operator at the core of its design.
In \cite{Tzanetos2023}, several algorithms are examined to discern whether their mechanisms produce desirable qualities in their respective domains. However, the primary findings of this study reject this hypothesis, as recent algorithms do not consistently align with the behavior or phenomenon on which they are based.

Another way to identify weak proposals is through experiments that unfairly enthrone algorithms with non-desired behaviors. An example is the work in \cite{Kudela2023}, which reveals the center bias present in many bioinspired algorithms by analyzing more than 100 bioinspired and evolutionary algorithms. The proposed methodology is to conduct experimental studies on functions whose optimal solution is situated at the center of the feasible set of solutions, and modifying them with specific displacements to avoid that. Under this methodology, it lists several algorithms that should not be used due to their strong center bias, including well-known proposals such as the Grey Wolf Optimizer, Dragonfly Optimization, Whale Optimization Algorithm, and Bird Swarm Algorithm. This work also recommends avoiding other algorithms for their lack of algorithmic novelty, including Harmony Search and Cuckoo Search, among others.
\begin{changed}
Another interesting work is \cite{rajwarStructuralBiasMetaheuristic2025} which, besides classifying different types of bias, proposes several techniques to detect structural bias in the exploration of an evolutionary algorithm. In addition, it provides an extensive list of different algorithms with a strong structural bias identified in the literature. The most common bias is toward the center, but there are also biases towards the boundaries and/or edges of the domain search, among others.
\end{changed}

\subsection{Pathways to Detect Weak Proposals} \label{sec:pathways_weak}

Through these reviews, it becomes evident that there is a growing interest in adequately comparing new proposals with existing ones in the literature. Based on previous examples, we have examined the methodologies used for this type of \textit{straw-grain} discrimination studies, and we propose pathways that researchers can embrace to enforce a greater rigor in the proposal of new algorithms, detecting their lack of novelty:
\begin{itemize}[leftmargin=*]
\item \textit{Equation-level equivalence}, either at the operator or at the component level. This method involves comparing the algorithmic and mathematical descriptions of the search operators of various algorithms towards detecting a possible lack of diversity among them. Techniques for this first pathway include: 
\begin{itemize}[leftmargin=*]
\item Conducting homologous component studies by analyzing general expressions of previously established algorithms to identify similarities and differences in their structure and behavior \cite{Hu2024}.

\item Utilizing formal verification methods to rigorously evaluate algorithmic redundancy \cite{urban2021}. The process begins by formally representing the algorithms to model their structure and behavior. Then, automated tools such as model checkers or theorem provers are employed to identify equivalences or redundancies by systematically comparing algorithmic components and behaviors.

\item Determining equivalence through operator simplification. This case implies analyzing whether certain components or operators in a bioinspired algorithm contribute substantially to its performance. To assess the impact of individual components, ablation tests should be employed. These tests systematically remove or modify parts of the algorithm to evaluate their specific contribution to the search performance (or to any other quality aspect that is relevant for the problem(s) at hand). 

\item Examining the standard definitions of operators to ensure that differences in terminology are not mistakenly interpreted as evidence of differing algorithmic behavior. By systematically analyzing and standardizing operator definitions, the community can more effectively compare meta-heuristic algorithms \cite{swan2015research,Swan2022}. In addition, it can be applied to compare different implementations of the same algorithm across different software libraries. 
\end{itemize}

\item \textit{Configuration-level equivalence}. This pathway refers to the equivalence between two algorithms when specific values of the parameters controlling their search operators can cause them to behave in a very similar manner. The challenge lies in determining when the behavior of one algorithm can be reduced to or generalized by another. Several results can produce different scenarios:
\begin{itemize}[leftmargin=*]
\item Non-innovative algorithm: If the performance of an acclaimed novel solver is essentially identical to an existing algorithm under a certain set of parameter values.

\item Generalization of an existing solver: If the new bioinspired solver introduces additional flexibility in its parameters or expands its search capabilities while maintaining equivalence with an earlier algorithm for specific parametric settings, it must be considered a more general version of the original algorithm,
making the original method redundant.

\item Genuinely innovative algorithm: If the novel algorithm introduces significant differences in performance or behavior that cannot be replicated by adjusting the parameter values of an existing algorithm.
\end{itemize}
\end{itemize}


The pathways outlined above allow for the identification of weak proposals based exclusively on their definition. However, the final judgment of the worth of an algorithm must also consider the experimental comparison that demonstrates its practical value. In the following section, we introduce methodologies to guarantee reliable experimental comparisons between bioinspired algorithms.

\section{Fair and Right Comparisons in Bioinspired and Evolutionary Optimization: Replicability and Benchmarks} \label{sec:frc}

To ensure that a proposal for a new bioinspired algorithm advances in the field with originality and practical impact, it is crucial to establish a well-defined experimental section that enables an impartial evaluation of its performance. This is typically associated with improving (or at least matching) the performance of state-of-the-art algorithms. However, the potential contribution of a new proposal is not necessarily limited to an improvement of its search performance (convergence speed or quality of solutions) with respect to the considered counterparts in the benchmark.

In Section \ref{sec:cons-real-world}, we outline several considerations that must be made to ensure that the final aim of bioinspired and evolutionary computation is to successfully solve real-world problems. In Section \ref{sec:pathways-ensure-fair}, we propose several pathways to guarantee fair and right comparisons. In Section \ref{sec:repl-exper-studies}, we underscore the importance of replicability in experimentation. Finally, Section \ref{sec:proper-benchmarks} emphasizes the importance of using appropriate benchmarks.


\subsection{Considerations in Real-World Problems}\label{sec:cons-real-world}

When using optimizers to address real-world problems, certain factors are of particular importance. In this regard, \citet{OSABA2021} outline the distinct requirements at various stages of the process, including design, development, experimentation, and deployment, to effectively address real-world optimization problems. They identified multiple challenges inherent to the design of new bioinspired algorithms and proposed a series of steps to be followed throughout the algorithm's development. These steps range from the initial phase of problem modeling to the final validation of the algorithm:
\begin{itemize}[leftmargin=*]
    \item \textit{Problem modeling and mathematical formulation}: It is recommended to begin with a thorough understanding of the real-world context and conceptualization of the problem, which should end in a formal mathematical formulation of the objectives, decision variables, and constraints involved in the problem. This mathematical formulation is essential for determining the structure of the problem and guiding the selection or development of suitable search algorithms. An important outcome of this step is a list of collected functional and non-functional requirements for the solver to be developed, which must be considered towards the design or selection of the solver and accounted for during the rest of the steps.
    \item \textit{Algorithmic design, solution encoding and search operators}: Once the problem is clearly defined, the next step is to design an algorithm that can efficiently search the solution space. This involves determining how to encode potential solutions, which is critical in bioinspired algorithms, as it directly affects how the algorithm explores and exploits the search space. In addition, it is important to ensure that the algorithm design aligns with the specific constraints and requirements of the problem being solved. This step may involve adapting existing metaheuristics or developing new search operators that better fit the problem's characteristics. The design must consider the functional and non-functional requirements collected during the first step of the process, including factors such as speed, simplicity, deployability in specialized software/hardware, or the possibility of paralleling the implementation of the designed search operators, among others.
    \item \textit{Performance assessment, comparison and reproducibility}: This phase is dedicated to evaluating the effectiveness of the proposed algorithm. The performance of the algorithm should be rigorously assessed using standard benchmarks or real-world data to determine if it meets the desired requirements. Comparisons with other state-of-the-art algorithms help contextualize their performance. Furthermore, replicability is a crucial factor, ensuring that the results are consistent and can be reproduced in different contexts. This assessment should also account for the specific needs and priorities of the real-world problem, such as computational efficiency, scalability, and robustness, which may be just as important as the raw performance of the optimization algorithm.
    \item \textit{Algorithmic deployment for real-world applications}: The final step involves implementing the algorithm in the real-world scenario. This phase tests the algorithm's ability to deliver practical solutions under actual operational conditions, which may differ significantly from the observed and modeled experimental environments. Issues such as scalability, robustness, and ease of integration into existing systems become paramount, considering their computational cost and adaptability to dynamic real-world conditions. It is at this step that the algorithm proves its true value for practical impact.
\end{itemize}

Although the above process can summarize the overarching goal of optimization research (\emph{to efficiently and effectively solve real-world optimization process}), bioinspired and evolutionary optimization research often assumes that the research objective is to design an algorithm that performs better than the state of the art on a certain set of benchmark problems. Although this is a reasonable target for early stage research aimed at producing innovative proposals, it should be subject to the same methodological standards in terms of experimentation and comparison between algorithmic choices. 


\subsection{Pathways to ensure Fair and Right Experimental Comparisons}\label{sec:pathways-ensure-fair}


The community is increasingly aware of the significant flaws exhibited by many new proposals, particularly in the experimental phase of their research. Researchers, like \citet{LATORRE2021}, have proposed methodological approaches to support more solid and principled experimentation in bioinspired and evolutionary computation. In this specific work, such guidelines can be summarized as follows:
\begin{itemize}[leftmargin=*]
\item \textit{Choice of benchmarks and algorithms for comparison}: Benchmarks are a fundamental aspect of algorithm evaluation. As discussed in Section \ref{sec:poor_exp_studies}, custom optimization functions can produce misleading conclusions.
  A more reliable analysis could be achieved by switching to a better, more standard benchmark or expanding to a larger set of problems \cite{Delser2021b}.
\citet{LATORRE2021} advocate for the use of well-designed standard benchmarks that include diverse test functions, representing a broad range of real-world problems. Comparisons with state-of-the-art algorithms should extend beyond just solution quality, incorporating aspects such as efficiency, simplicity, and convergence speed.
  This kind of analysis can guide future research and improvements in the algorithm's design.
    
\item \textit{Validation of the results}: In the context of metaheuristic algorithms, it is insufficient to present only the raw results of the algorithm's performance; statistical validation is a critical step of the validation of such results. Meta-heuristic algorithms, by their nature, are stochastic and produce variable results, making statistical tests \cite{derrac2011practical,carrasco2020recent} essential to determine whether observed differences in performance are statistically significant or simply due to the stochastic nature of the operators inside the solver. Fortunately, statistical validation has become a standard practice in the meta-heuristics literature. Additionally, visualization techniques are extremely valuable, as they help condense complex experimental information, even during the run of algorithms, into more interpretable forms that support and favor the quick assessment of the algorithm's behavior.
    
\item \textit{Components analysis and parameter tuning of the proposal}: The hypotheses of the proposal should be clearly outlined at the beginning of the paper and revisited during the validation of results. Furthermore, authors should conduct a comprehensive analysis of results, covering key aspects such as the search phase, the component analysis, the algorithm parameter tuning, and the statistical comparison with state-of-the-art algorithms.
This comprehensive examination guarantees a robust evaluation of the proposed method and its performance in comparison to existing approaches.
    
\item \textit{Why is my algorithm useful?}:
Prospective contributors should provide a clear and detailed explanation of why their proposed algorithm is worthy of attention and consideration within the research community. They should demonstrate that their algorithm is competitive with existing, state-of-the-art methods or present a compelling case for the methodological contributions that their work makes to the field. This clarity of communication is essential for conveying the significance of the proposed algorithm and its potential impact on the field.
\end{itemize}

In recent times, the community has taken up this role by proactively proposing different methodologies, tools, and frameworks. One of the most relevant directions followed in this regard is the proposal of standardized benchmarks and experimental conditions. An important advantage of them is that they facilitate algorithm comparison since researchers can directly compare their results in a transparent, externally verifiable, and replicable fashion. 

\subsection{Pathways to support the Replicability of Experimental Studies}
\label{sec:repl-exper-studies}

This is particularly relevant due to the questionable reproducibility or replicability of the results reported for many bioinspired algorithms 
in the literature. Replicability is one of the motivations at the heart of the framework proposed in \cite{Swan2022} to support the verifiable comparison of new meta-heuristic approaches. This framework hinges on several components:
\begin{itemize}[leftmargin=*]
\item \textit{Extensible and re-usable templates}: These templates are designed to offer flexibility by allowing researchers to configure their behavior through an extensible palette of components. This modular approach eliminates the need for altering existing code, making it easier to adapt, extend, and repurpose algorithmic components for a variety of problems.
\item \textit{White-box problem descriptions}: A description of the problem with analytic information that could be used to guide the algorithm's selection or construction. Providing detailed transparent descriptions of optimization problems, including analytic information such as objective landscape characteristics or known constraints, can guide the selection or design of operators that are well-suited to the specific problem under consideration.

\item \textit{Remotely accessible frameworks, components, and problems}: Hosting algorithm frameworks, optimization components, and a curated set of benchmark problems on a remotely accessible platform can foster reproducibility and collaborative research.
Researchers could directly access them, enabling widespread reuse, replicability, and shared knowledge discovery.

\end{itemize}

\subsection{Pathways to guarantee Proper Benchmarks}\label{sec:proper-benchmarks}

The limited variety of test functions commonly used in benchmarks within the community is another issue that has garnered research attention. A recent advancement in this direction is presented in \cite{Sharma2024}, which provides extensive information on practical scenarios for the development of novel optimization algorithms. The review covers more than two hundred mathematical functions and more than fifty real-world engineering design problems. Recognizing the critical need for robust experimental evaluations to accompany the design of bioinspired algorithms, this work offers a comprehensive array of options for assessing the quality and effectiveness of new proposals.

Concerns about design biases in optimization research
have increasingly drawn attention in recent years. As highlighted in \cite{Kudela2022}, the center bias can distort the results and misrepresent the true performance of algorithms. To address this issue, \citet{Kudela2023} proposed a systematic procedure to identify methods that exhibit such biases, accompanied by a comprehensive analysis of bioinspired and evolutionary algorithms up to 2022. Although the adoption of diverse and well-designed benchmark functions, such as those reviewed in \cite{Sharma2024}, can mitigate the occurrence of center bias, other forms of bias may still persist. This underscores the critical need for new methodologies to effectively detect and counteract these biases in algorithmic evaluations, ensuring fairness and rigor in optimization research.

In response to this need, \citet{Walden2024} propose a novel mechanism based on statistical tests to identify potential biases in algorithms. Their benchmark comprises two identical and symmetrically located global optima, one positioned at the origin. By employing non-parametric statistical tests to analyze the best solutions across independent runs, the authors can detect significant behavioral differences around the optima. If such differences exist, the algorithm under evaluation is flagged as potentially origin-biased. Notably, their study identifies several algorithms that fail this test, including well-known algorithms.
This work not only highlights specific instances of bias but also sets a precedent for systematic bias detection, paving the way for more reliable evaluations of bioinspired and evolutionary solvers.

As a final contribution to the advancement in this field, it is essential to ensure that algorithms are effective across a wide range of problems.
In this context, \citet{Kumar2021,Kumar2020} have developed two benchmarks that cover nearly 50 diverse problems, aimed at facilitating more meaningful comparisons between algorithms and providing insights into their performance in real-world scenarios. Furthermore, the researchers conducted a comprehensive comparative analysis of state-of-the-art algorithms, culminating in a ranking system that serves as a valuable tool for evaluating the efficacy of new algorithmic proposals. 

\section{Methodologies and Pathways for Improving Existing Bioinspired Optimization Algorithms}\label{sec:methodologies_improving}



In the comprehensive review by \citet{Velasco2024}, it is revealed that 65\% of the studies analyzed focus on improved versions of established algorithms, signaling a shift away from creating bioinspired algorithms based on novel analogies.
This review examines over a hundred recent studies containing terms such as ``new'', ``hybrid'', or ``improved optimization'' in their titles. The vast majority of these proposals are actually refinements of existing solvers.

Unfortunately, the study in \cite{Velasco2024} underscores a relevant issue: most of these algorithms address engineering problems that lack direct applicability to real-world scenarios.
Addressing this gap is the focus of the current section: to explore potential pathways for enhancing an existing bioinspired solver, making it more suitable for addressing specific problems or families of problems with greater relevance and utility. Such pathways are enumerated and described in what follows:
\begin{itemize}[leftmargin=*]
\item \textit{Automatic Equivalence between algorithms}: We begin with a common methodological approach, to add mechanisms or change existing ones for improved versions, thereby increasing the complexity of the model. When this is the case,
ablation tests are highly recommended to assess the relative contribution of such added mechanisms and/or modifications to the acclaimed performance improvement.

Given that ablation tests can be very time-consuming, several methodologies have recently emerged to automate this process.
Among them, we highlight \cite{Fister2021}, which discusses how to automatically detect equivalence between stochastic population-based nature-inspired algorithms considering both a conceptual level -- using the objective function value -- and an operational level, expressed in terms of a measure of the population diversity.
Their notion of equivalence requires that the average fitness values and the diversity of the population must not be significantly different in each generation for two algorithms under comparison to be equivalent to each other. Through the use of Markov chains as a representation of population and fitness, this methodology can assess the similarity of bio-inspired algorithms.

\item \textit{Decomposition into their canonical components}: 
  This is the approach followed in \cite{DeArmas2022}, which presented a methodology for the decomposition of bioinspired algorithms. Specifically, the methodology considers a set of templates that act as a framework to decompose and analyze bioinspired algorithms based on their characteristics and in a structured manner.
In this pool of templates, we find different categories of components, such as the method of generating initial solutions, the solution pool itself, the solution archive, update mechanisms, input and output functions of the solution pool, and others. With all this, a similarity analysis between bioinspired algorithms can be performed by comparing their components within the pool of templates, facilitating not only the evaluation of algorithmic equivalence between them, but also a substrate of components for the automated adaptation and/or construction of meta-heuristic algorithms \cite{stutzle2019automated,zhao2023automated}.

\item \textit{Evaluation of whether a metaheuristic algorithm is derived from an existing one}: \citet{Hu2024} have recently introduced a novel methodology to evaluate the originality and research aim of bioinspired algorithms. This approach aims to discern whether a metaheuristic algorithm is a derivation of an existing method, based on the mathematical characterization of its compounding operators. Central to this methodology is the distinction between \emph{root algorithm} and \emph{homologous algorithm}, categorized by their search operators. A \emph{root algorithm} represents a genuine innovation in the design of search operators, whereas a \textit{homologous algorithm} is recognized for its practical utility or academic significance, even if it does not introduce fundamental algorithmic changes. The methodology proposed in this work defines two key constructs: perturbation mapping and difference mapping, which serve as formal representations of search operators. If a given operator of a bioinspired algorithm can be expressed as a simple (especially linear) combination of these mappings derived from a \emph{root algorithm}, the new algorithm is classified as \emph{homologous}. This distinction allows researchers to determine whether an adaptation of a metaheuristic algorithm represents a true contribution or is an incremental variation without substantive novelty. To demonstrate its utility, the authors have applied this framework to several widely used algorithms, comparing them with their root counterparts. 

\item \textit{Mathematically proven convergence improvements}: 
\citet{Choi2023} introduces a mechanism to ensure the global convergence of swarm intelligence algorithms. This method involves applying stochastic perturbations to half of the swarm agents, followed by projecting all agents back into the feasible solution space to respect problem constraints. This projection is particularly critical in optimization problems with bounded search spaces. By incorporating controlled randomness, the perturbation mechanism helps maintain diversity within the swarm and prevents premature convergence by encouraging exploration in overly exploited areas. Remarkably, \cite{Choi2023} provides mathematical proof that this approach, when integrated into a broad class of swarm algorithms, ensures global convergence under specific conditions.

\item \textit{Dynamic recalibration of metaheuristic algorithms}: \citet{Jia2024} proposes a novel update mechanism designed to dynamically recalibrate metaheuristic algorithms. The mechanism adjusts the balance between exploration and exploitation based on the current population distribution. It employs statistical analysis, calculating the standard deviation of agents' historical positions over recent generations, to measure population dispersion. This information is used to drive the algorithm, iteratively adapting its search behavior. The versatility and efficacy of this adaptation strategy are demonstrated with more than 50 different algorithms, demonstrating significant performance improvements across a diversity of solvers. 

\end{itemize}

\section{A Smart and Promising Solution for overcoming Known Critical Issues: Automated Design of Metaheuristic Algorithms}
\label{sec:automation_design}

Advancing towards the automated design of metaheuristic algorithms has marked a significant leap in bioinspired and evolutionary optimization. This involves refining individual algorithms and automating their creation.
The automated design of metaheuristic algorithms can provide breakthroughs in various scientific and engineering domains, making the automated design of metaheuristic algorithms the norm when addressing real-world optimization problems.

In this section, we describe works that tackle this automated design process. First, in Section \ref{sec:autom-design-metah}, we revisit several notable proposals to optimize the design. Next, Section \ref{sec:using-large-language} pauses at several recent contributions that incorporate Large Language Models (LLMs) for the design of new metaheuristic algorithms.

\subsection{Automating the Design of Metaheuristic Algorithms}
\label{sec:autom-design-metah}

\citet{Camacho2023} explore whether manually designing or automating the construction of metaheuristic algorithms can be a superior approach, yielding the following main observations:
\begin{itemize}[leftmargin=*]
\item The manual design process often involves seeking inspiration from other fields of knowledge and iteratively refining algorithms based on trial and error. This approach, while effective, can be resource-intensive and time-consuming, as it requires substantial human involvement in identifying the optimal algorithmic design from a multitude of possibilities. 
\item In contrast, the automated design approach emphasizes the elimination of human intervention by leveraging recent advances in automatic algorithm configuration techniques. These techniques include hyper-heuristics, evolutionary programming, and reinforcement learning, which operate over the design space defined by algorithmic operators.
\end{itemize}

As further detailed in \citet{zhao2023automated}, automated design methods can efficiently explore a vast space of algorithmic components, automatically selecting and combining operators in a way that optimizes performance. This not only accelerates the development of metaheuristic solvers but also offers the potential to discover novel algorithmic structures that may not be easily identified through manual design. By automating the creation process, researchers can focus more on 
applying these algorithms to real-world problems.

In general, the automatic development of metaheuristic algorithms involves two key steps:
\begin{enumerate}[leftmargin=*]
\item 
The use of an automatic configuration tool to identify the optimal composition of operators and processing directives for a given algorithm. In this phase, the algorithm's performance is evaluated across various problem instances, and its configuration is adjusted iteratively until a predefined computational budget is met. Once this process is complete, the algorithm is considered to be fine-tuned with the best possible configuration. 
    
\item The second step focuses on enabling the algorithm to be used for comparisons across different problem instances and configurations. To achieve this, a metaheuristic software framework is employed to define and implement the design space of the metaheuristic. This framework provides the infrastructure necessary to systematically explore various algorithmic components and their interactions. Subsequently, the framework is iterated with the previous step to automatically generate a fully functional algorithm capable of being tested against other problem configurations. 
\end{enumerate}

The development of software tools capable of supporting the creation, automatic generation, and evaluation of new algorithms represents one of the most promising advances in metaheuristic optimization research. Such tools can evaluate potential configurations and identify the best fit for a given problem. In \cite{Acosta2024}, the authors introduce a novel software framework for creating new algorithms. Their software is built upon a standardized theoretical framework for analyzing, modifying, and generating metaheuristic algorithms. It features three distinct panels, each dedicated to a specific functionality. These panels enable users to configure continuous optimization problems and assemble new algorithms in a modular and code-free manner. The software framework provides access to over 110 benchmarking domains and includes robust statistical analysis, reporting key metrics such as fitness and runtime, alongside convergence plots to track algorithm progress. Tools alike, in general, can help boost and democratize the development of new proposals for optimization algorithms.


\subsection{Using Large Language Models to automate the Design of Metaheuristic Algorithms}
\label{sec:using-large-language}

Large Language Models \cite{Mikolov2013} are artificial intelligence systems trained on vast amounts of data and are capable of processing and generating human-like text. LLMs can summarize texts, answer questions, translate languages, and even generate creative content. Recently, LLMs have been used to design and/or modify different algorithms. Thus, they can be considered an important and promising approach to automate the design of new evolutionary solvers. We herein comment on several recent examples of this trend.

A first example of this automatic approach is presented in \citet{Vanstein2024}, where evolutionary Large Language Models (LLaMEA) are utilized to generate novel algorithms. Given a set of criteria and a task definition (search space), LLaMEA iteratively generates, mutates, and selects algorithms based on performance metrics and feedback from runtime evaluations. This methodology takes advantage of the capabilities of LLMs to create new metaheuristic algorithms that can then be evaluated and compared across a wide range of problem instances. 

Another approach that harnesses contextual knowledge embedded in LLMs is FunSearch \cite{romera2024mathematical}, which has represented a breakthrough in automated discovery of programs, hence having the potential to be applicable to generate and evolve metaheuristic solvers. By combining an LLM with an evolutionary framework, FunSearch systematically explores algorithmic design spaces, which can create novel optimization techniques that autonomously improve their performance through iterative refinement and runtime evaluation. Unlike traditional manual design, FunSearch can enable the discovery of unexpected and potentially superior bioinspired algorithms across different problem domains. 

\citet{liu2024example} have recently introduced the so-called Algorithm Evolution using the LLM (AEL) approach to designing optimization algorithms, with a specific focus on guided local search techniques. This framework leverages an LLM to generate evolutionary search operations, including initialization, crossover, and mutation, enabling the automated evolution of novel algorithmic solutions. AEL begins by prompting the LLM with a query that specifies the algorithm design task, information about the parent algorithms, and the expected format of the output. By harnessing the generative capabilities of the LLM, AEL 
reduces human involvement, and enables the discovery of new optimization methods.

The automated design of metaheuristic algorithms leveraging the rich knowledge embedded in LLMs represents a promising avenue for addressing the long-standing issue of solvers drawing inspiration from arbitrary or meaningless biological metaphors. However, to fully realize this potential, our pathways identified in previous sections are still required to ensure that the automated design of metaheuristic algorithms not only overcomes existing limitations, but also provides tangible practical value. 


\section{Conclusion and Outlook} \label{sec:conclusions}

\begin{figure}[h!]
    \centering
    \includegraphics[width=\linewidth]{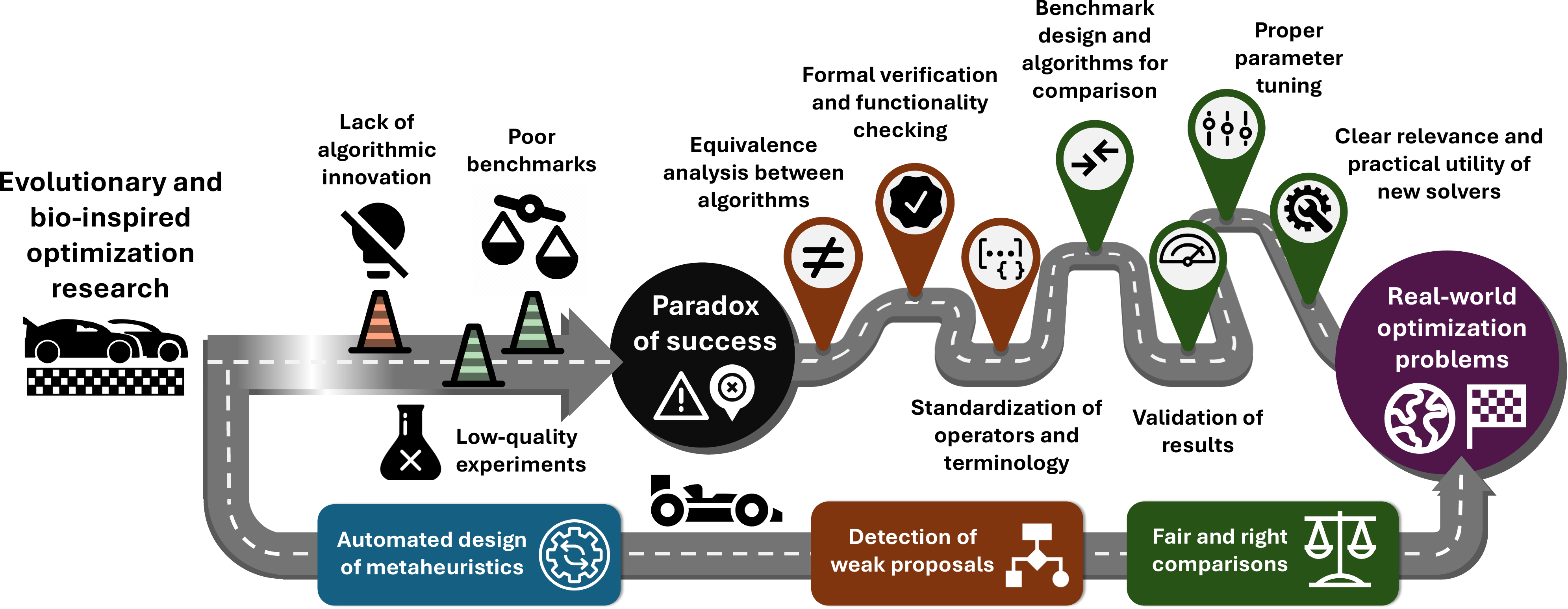}
    \caption{Graphical summary of the issues, pathways and promising directions discussed in this manuscript.}
    \label{fig:summary}
\end{figure}

In this paper, which we graphically summarize in Figure \ref{fig:summary}, we have critically examined challenges in evolutionary and bioinspired optimization, and outlined methodological ways to address them effectively. We have emphasized the need to differentiate truly innovative contributions (\emph{the wheat in the chaff}), and have highlighted the importance of robust, non-biased experimental practices. Through recent examples in the literature, we have showcased strategies for the design, automated construction, improvement, and evaluation of bioinspired solvers, ensuring a more transparent assessment of the components of new proposals, and their fairer and more solid comparison to established metaheuristic methods.

The analyzed pathways are designed to refocus optimization research on its ultimate goal: developing algorithms capable of efficiently solving real-world problems. We strongly encourage the research community to adopt these recommendations as essential guidelines for achieving this objective. Researchers, editors, and reviewers must actively integrate the methodologies and best practices highlighted in this work to cultivate a more rigorous and innovative environment in evolutionary and bioinspired optimization. 

Although the field has made significant progress, it still faces a long journey to realign itself and produce optimization and search tools capable of addressing major scientific and technological challenges across diverse disciplines. Ultimately, embracing these guidelines will significantly contribute to the advancement of knowledge and to the development of cutting-edge solutions to address real-world challenges and to redirect the field towards new metaheuristic solvers that deliver real practical value.


 
\section*{Acknowledgments}

This work is supported by the Knowledge Generation Projects PID2023-149128NB-I00 and PID2023-150070NB-I00, funded by the Ministry of Science, Innovation, and Universities of Spain. J. Del Ser acknowledges funding support from the Basque Government through the consolidated research group MATHMODE (ref. IT-1456-22).

\section*{CRediT Author Statement}

\textbf{Daniel Molina}: Conceptualization, Methodology, Investigation, Writing Original Draft, Writing - Review \& Editing, Supervision, Project administration; \textbf{Javier Del Ser}: Conceptualization, Methodology,  Writing - Review \& Editing, Supervision, Project administration; \textbf{Javier Poyatos}: Conceptualization, Methodology, Investigation, Writing - Original Draft; \textbf{Francisco Herrera}: Conceptualization, Methodology, Writing - Review \& Editing, Supervision, Project administration.

\section*{Declaration of AI-assisted technologies in the writing process}

During the preparation of this work, the authors used large-language models to improve the readability and language of the manuscript. After using this tool/service, the authors reviewed and edited the content as needed and assumed full responsibility for the content of the published article.

\bibliographystyle{elsarticle-num-names}
\bibliography{paradoxEC_paper}

\end{document}